\documentclass[runningheads]{llncs}
\usepackage[T1]{fontenc}
\usepackage{booktabs} 
\usepackage{multirow} 
\usepackage{graphicx,pifont}
\usepackage{amsmath,amssymb} 

\DeclareMathOperator*{\argmin}{arg\,min}
\begin{document}
\title{Self-supervised Multi-actor Social Activity Understanding in Streaming Videos}
%
%
\author{Shubham Trehan\inst{1} \and
Sathyanarayanan N. Aakur\inst{1}}
\authorrunning{S. Trehan et al.}
%
\institute{CSSE Department, Auburn University, Auburn, AL, 36849
\email{\{szt0113,san0028\}@auburn.edu}}
\maketitle              
\begin{abstract}
This work addresses the problem of Social Activity Recognition (SAR), a critical component in real-world tasks like surveillance and assistive robotics. Unlike traditional event understanding approaches, SAR necessitates modeling individual actors' appearance and motions and contextualizing them within their social interactions. Traditional action localization methods fall short due to their single-actor, single-action assumption. Previous SAR research has relied heavily on densely annotated data, but privacy concerns limit their applicability in real-world settings. In this work, we propose a self-supervised approach based on multi-actor predictive learning for SAR in streaming videos. Using a visual-semantic graph structure, we model social interactions, enabling relational reasoning for robust performance with minimal labeled data. The proposed framework achieves competitive performance on standard group activity recognition benchmarks. Evaluation on three publicly available action localization benchmarks demonstrates its generalizability to arbitrary action localization. 

\end{abstract}
\section{Introduction}
Social activity recognition (SAR) is a key part of computer vision applications in the real world, such as surveillance and assistive robotic systems. 
It differs from traditional event understanding approaches~\cite{soomro2015action,soomro2017unsupervised,aakur2022actor,aakur2020action,carreira2017quo} since it requires the modeling of individual actor's appearance and their motions, and contextualizing them within the scope of their social interactions. 
SAR brings a unique set of challenges. 
First, there is a need for actor localization, social relationship modeling, and social activity recognition. Second, the number of actors in each frame can change due to occlusion, camera range, or notice due to missed/false detection. Finally, a scene can have an arbitrary number of social groups. Traditional action localization approaches~\cite{aakur2020action,aakur2022actor,soomro2017unsupervised} cannot be directly extended to this problem since they assume a single action performed by a single actor. 

The dominant approach has been to learn the social dynamics of a scene using attention-based or graph-based relational reasoning in a supervised learning setting. The key assumption has been the availability of densely annotated data for training and near-perfect actor localization. Hence, the literature has focused on feature aggregation across time and social groups. While this has yielded tremendous progress, it is not always possible to expect densely annotated data for training, primarily due to the privacy concerns involved in collecting, storing, and annotating visual data in a social setting. There is a need to move away from over-reliance on labeled training data and towards self-supervised learning approaches that can learn in an open world, i.e., unconstrained training and test semantics, and in a streaming fashion, i.e., learning with a single pass through the data without storing it without loss of generalization. 

In this work, we focus on addressing social activity understanding in streaming videos without labeled data. We propose the idea of multi-actor predictive learning for jointly modeling actor-level actions and contextual, group-level activities. We move away from the single-actor, single-action assumption from prior approaches~\cite{aakur2019perceptual,aakur2020action,aakur2022actor,trehan2022towards} and propose to represent visual scenes in a social-contextualized action graph that provides an elegant, end-to-end trainable framework for social event understanding.
The \textbf{contributions} of our approach are four-fold: (i) we are the first to tackle the problem of self-supervised social activity detection in streaming videos, (ii) we introduce a visual-semantic graph structure called an \textit{action graph} to model the social interaction between actors in a group setting,  (iii) we show that relational reasoning over this graph structure by spatial and temporal graph smoothing can help learn the social structure of cluttered scenes in a self-supervised manner requiring only a single pass through the training data to achieve robust performance, and (iv) we show that the framework can generalize to arbitrary action localization without bells and whistles to achieve competitive performance on publicly available benchmarks.

\begin{figure}[t]
    \includegraphics[width=0.99\textwidth]{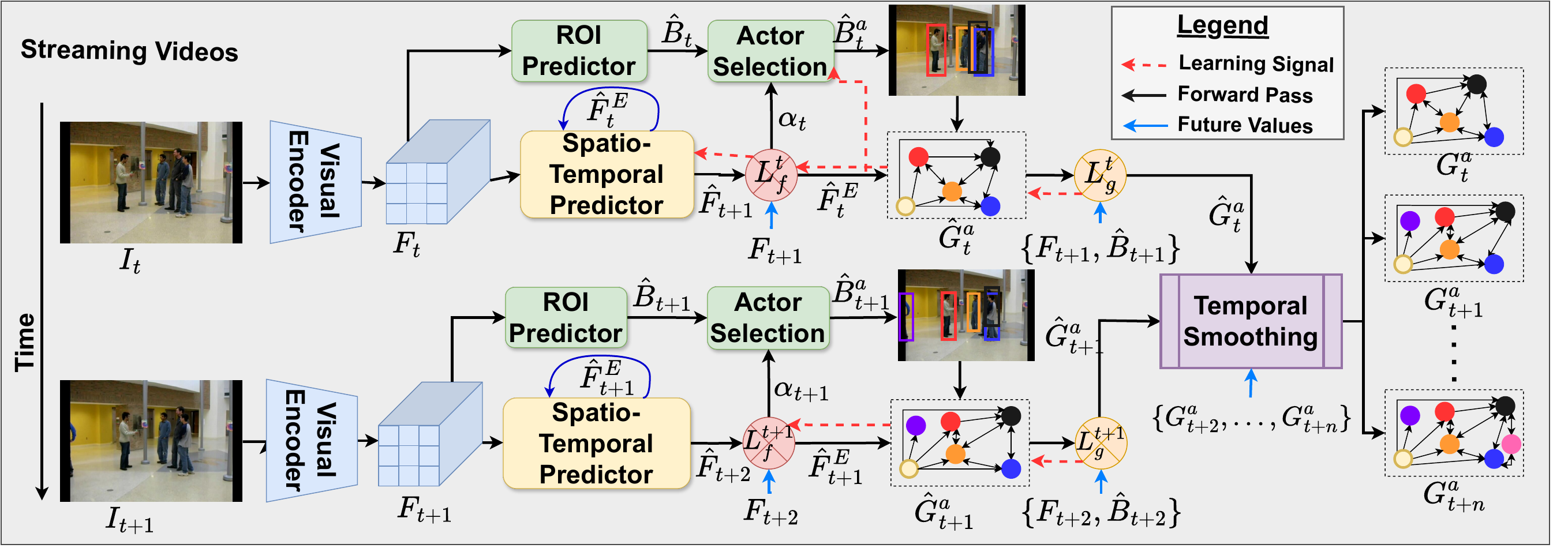}
    \caption{\textbf{Overall architecture}. Using multi-actor predictive learning, we can localize actors and model their interactions as an action graph, which can be used for downstream event understanding tasks such as action and social activity detection.}
    \label{fig:arch}
\end{figure}
\section{Related Work}
\textbf{Group activity recognition} has been a widely studied area of social event understanding. The typical pipeline starts with actor detection, individual feature extraction, and social interaction modeling. Action features are extracted for each actor using pre-trained action recognition models~\cite{carreira2017quo}. 
The primary mechanism has been to model the social interaction of actors within a group setting using attention-based mechanisms to generate social group features~\cite{tamura2022hunting}, model individual actor dynamics~\cite{gavrilyuk2020actor}, to model the spatial and temporal dependencies~\cite{li2021groupformer} jointly, or for multi-view representation learning~\cite{raviteja2023sogar}. 
Others use transformers~\cite{vaswani2017attention} to bypass object detection requirements~\cite{kim2022detector}, model keypoint dynamics~\cite{yuan2021learning}, social relation modeling~\cite{ehsanpour2020joint}, and spatiotemporal multiscale feature aggregation~\cite{zhou2022composer} to reduce training requirements. 

\textbf{Relational reasoning} is another line of work that focuses on aggregating actor-actor interactions for group activity understanding. These approaches aim to model the spatiotemporal dependencies by considering the spatial relationships between objects using a variety of mechanisms such as aggregating the relational contexts and scene information using transformers~\cite{pramono2020empowering}, using graph convolutional networks (GCNs) to capture the appearance and position relation between actors~\cite{wu2019learning}, or capture spatial coherence using recurrent neural networks (RNNs)~\cite{8621027,wang2017recurrent}, convolutional neural networks (CNNs)~\cite{azar2019convolutional}, graph-LSTMs~\cite{tang2019coherence}, graph attention~\cite{8769904}, factor graphs~\cite{xie2024active}, knowledge distillation~\cite{tang2019learning}, tokenization~\cite{wu2022active}, tracking~\cite{thilakarathne2024group}, and contrastive learning~\cite{han2022dual}, to name a few. The prevalent paradigm in the above approaches has been supervised learning to establish and learn social interactions, with varying levels of supervision, i.e., bounding box locations and labels of individual actors, group activity labels, and social group memberships, which requires immense human effort for annotations and may reduce their generalizability. 
Some works~\cite{yan2020social,kim2022detector,zhou2022composer,yuan2021learning} have attempted to reduce the training requirements by relaxing assumptions about the availability of annotations but still require a large amount of labeled data. 

Our work is one of the first to tackle this problem from a self-supervised learning perspective by modeling the actor dynamics from a multi-actor predictive learning perspective. 
\textit{Predictive learning} has emerged as a powerful paradigm for visual event understanding. Proposed in cognitive science literature~\cite{zacks2001perceiving}, the goal of predictive learning is to learn representations by anticipating the future and use the residuals for downstream tasks such as event segmentation~\cite{aakur2019perceptual}, action localization~\cite{aakur2020action,aakur2022actor}, active object tracking~\cite{trehan2022towards}, future frame generation~\cite{lotter2016deep}, and hierarchical event perception~\cite{mounir2024streamer}, among others. All prior works have focused on single-actor settings, where only one global action is expected to be present. We offer a unique perspective on predictive learning by extending the idea to multi-actor predictions for group activity detection. We do not require any annotations and aim to learn robust representations at both the actor level and group level, while feature aggregation allows us to model social interactions.  

\section{Proposed Framework}
\textbf{Overview.} We propose to tackle the problem of multi-actor, multi-action localization in streaming videos. 
The overall framework is illustrated in Figure~\ref{fig:arch}.
Given a sequence (stream) of video frames $\{I_0, I_1, \ldots, I_t\}$, we aim to localize actors of interests, characterized by their location (represented as bounding boxes) $\hat{B}^a_i$ and visual features $F^a_t$. We then construct a graph structure called an \textit{action graph} ($\hat{G}^a_t$) whose nodes are actors and edges are social interactions, along with an event node. A composite spatiotemporal graph is fed through a temporal smoothing layer that contextualizes event and action level features to construct a final action graph that can be used for various downstream tasks such as group activity understanding and social activity understanding and can be extended to arbitrary action localization. We present each module in detail below.  

\subsection{Visual Perception and ROI Prediction}\label{sec:vis_percep}
Our framework begins with a visual perception module, aiming to extract visual features at both scene and object levels. 
Our primary visual perception module uses a DETR~\cite{carion2020end} model. 
For every frame $I_t$, we extract (i) global scene features, $F_t$, (ii) object regions of interests (ROI), $\hat{B}_t$, and (iii) object-level features, $F^B_t$. The ResNet backbone provides a lower-resolution, global feature map $F_t \in \mathbb{R}^{2048\times H \times W}$, where $W$ and $H$ are the spatial resolution of the global feature map. DETR's detection heads are used to generate initial object ROI proposals $\hat{B}_t$, i.e., the search space for actor localization. and the decoder outputs for each ROI prediction are used for object-level features ($F^B_t$). Note that at this stage, we only generate actor candidates that will be refined using the actor selection module described in Section~\ref{sec:action_graph}. We do not fine-tune DETR on the video datasets and use a model pre-trained on MS-COCO~\cite{lin2014microsoft}. During training, all objects are considered as candidate actors in a class-agnostic manner, following prior works~\cite{aakur2020action,aakur2022actor}, while we filter out only ``human'' predictions during inference. This allows us to learn the dynamics between human and non-human actors during training while allowing us to focus only on social dynamics during inference. 

\subsection{Spatiotemporal Prediction for Actor Localization}\label{sec:predLearn}
We model the spatial-temporal dynamics of the scene using a spatiotemporal predictor module. The goal is to learn an event-level representation ($\hat{F}^E_t$) that captures how each object, represented by their location $B_t$ and visual features $F^B_t$, change over time. We use a simple $L$-layer LSTM stack as our spatiotemporal predictor, which takes as the global scene-level feature $F_t$ as input and anticipates the future global representation $F_{t+1}$. The goal is not to predict the future frame by pixel-level regression, but rather model how the scene changes over time. This event representation $\hat{F}^E_t$ is continuously updated at every time instant $t$ as new frames ($I_t$) are observed in a streaming fashion using a predictive loss function given by
\begin{equation}
    \mathcal{L}_{global} = \frac{1}{H*W} \sum_{H,W} M_t \odot \lvert\lvert F_{t+1} - \hat{F}_{t+1} \rvert\rvert_2
    \label{eqn:global_pred}
\end{equation}
where $M_t$ is the motion difference between frames $I_t$ and $I_{t+1}$ computed as the first-order hold between $F_t$ and $F_{t+1}$; $\hat{F}_{t+1}$ is the anticipated global feature at time $t{+}1$, obtained by projecting the event feature $\hat{F}^E_t$ back to the 2-D feature space. Hence, the predictive loss attempts to force the LSTM stack to learn a robust event representation ($\hat{F}^E_t$) that can anticipate the future scene's spatial features ${F}_{t+1}$. The overall prediction errors are summed and normalized based on the number of spatial features regressed. 

\textbf{Actor Selection}
The unnormalized prediction errors ($P_t{=}M_t \odot \lvert\lvert F_{t+1} - \hat{F}_{t+1} \rvert\rvert_2$) from Equation~\ref{eqn:global_pred} are proportional to the predictability of each spatial location. Hence, higher prediction errors indicate the presence of a less predictable, foreground action(s), while lower prediction errors indicate a more predictable, background action. We formulate a prediction-driven attention mask $\alpha_t$ by passing $P_t$ through a softmax activation function to increase focus on foreground actions while suppressing background actions. The top $K$ attention ``slots'' are used to filter object ROIs $\hat{B}_t$ and select the \textit{actor} ROIs $\hat{B}^a_t$. 
Note that actor ROIs $\hat{B}^a_t$ are predicted only if the prediction-based attention slots $\alpha^{ij}_t$ fall within any ROI $b^k_t \in \hat{B}_t$. Hence, the number of actors chosen from the list of candidate ROIs is much lower, allowing us to model actor-level dynamics better using action graphs, as described next. 

\subsubsection{Building an Action Graph}\label{sec:action_graph}
To model actor-level interaction dynamics, we construct a graph $\hat{G}^a_t$ for every observed frame $I_t$, with actors as nodes $\mathcal{V}_t$. Each node $N_i \in \hat{G}^a_t$ is described by a feature vector $_i\hat{F}^a_t{=}[_iF^B_t;{ }_i\hat{B}^a_t]$ that captures its geometry and visual features. The edges in this graph structure, $\mathcal{E}_t$, are defined by the spatial structure of the actors selected using the prediction-based attention $\alpha_t$. Unlike previous graph-based approaches~\cite{ehsanpour2020joint,wu2019learning}, we do not use a fully connected structure. 
Instead, we model their social connectivity using a distance-based formulation. 
An adjacency matrix $A_t$ is constructed by computing the spatial proximity between each pair of nodes, given by the Euclidean distance $\phi$ between their locations and spatial geometry, and centering it by subtracting the mean distance between all nodes. The adjacency for each node $N_i$ is normalized as $A^i_t{=}\sigma(\frac{A^i_t}{\lvert\lvert A^i_t \rvert\rvert_2})$, to ensure that the distances are scaled proportionally and $\sigma$ is the Sigmoid function. The adjacency matrix is thresholded to get the final social structure by discarding all edges less than the average normalized distance in the adjacency. This formulation allows us to model the social interactions between the actors detected in the scene without the underlying assumption that all actors interact with each other, regardless of their social activity. Finally, an ``action node'', instantiated by the event features $\hat{F}^E_t$, is added to the graph and is connected to all actor nodes. This additional node allows us to propagate action features to relevant actors and the connections between actor nodes will enable us to capture contextual cues for modeling actions with interacting actors, as described in the next section. Empirically, in Section~\ref{sec:results}, we see that adding the action node and the subsequent contextualization using graph and temporal smoothing plays a big role in improving the performance of both group activity recognition and individual activity detection. The action graph formulation distinguishes us from prior unsupervised event understanding approaches~\cite{aakur2019perceptual,aakur2020action,aakur2022actor} since it allows us to model each actor individually without any prior assumptions about their role or interactions in a social group setting. 

\subsection{Contextualizing Cues with Graph and Temporal Smoothing}\label{sec:smoothing}
Recognizing social activity and individual actions in a group setting requires reasoning over the spatial interaction between actors at every instant and its evolution over time. To this end, given our action graph $\hat{G}^a_t$, the next step is contextualizing each person's action using a two-step spatial-temporal graph smoothing process. First, we use a message passing layer, as introduced in Graph Convolutional Networks (GCNs)~\cite{wu2019learning}, for spatial reasoning over the social interaction between actors as captured in $\hat{G}^a_t$. Formally, this is defined as 
\begin{equation}
    {F}^a_t = \sigma(A_t \hat{F}^a_t W_s)
    \label{eqn:spatial_msgPassing}
\end{equation}
where $A_t$ is the adjacency matrix for $G^a_t$, $\hat{F}^a_t$ is the feature representation for each node of the action graph, $W_s$ is the learnable parameter matrix for the GCN layer, and $\sigma$ is the ReLU activation function. The resulting features ${F}^a_t$ are contextualized across actors, conditioned on their social structure (specified by weighted edges $\mathcal{E}_t$), and the event-level features $\hat{F}^E_t$ represented by the action node in $G^a_t$. While this reasoning layer can be repeated, additional layers harm the model's performance (see Section~\ref{sec:results}) due to the homogenization of features. 

For temporal contextualization, we construct a composite spatial-temporal graph by establishing temporal edges between actor nodes in $G^a_t$ with their corresponding nodes in the subsequent graph $G^a_{t+1}$. 
While straightforward in theory, we must address two critical challenges for implementation. First, we do not have the ground truth tracking annotations that would enable us to establish actor-actor correspondences across frames. 
Second, the number of detected actors is not constant across time, requiring comparing graphs of different sizes. Hence, registering nodes across actor graphs between consecutive frames requires us (i) to establish a permutation matrix $\mathcal{P}$ to account for varying node ordering across graphs and (ii) to add null nodes (representing missed/false detections) to the graph with fewer nodes to ensure every node is registered to one node across time. The optimal permutation matrix $\mathcal{P}$ is obtained by computing a one-to-one match between two graphs $G^a_t$ and $G^a_{t+1}$ using the Hungarian matching algorithm to minimize the distance between the two graphs. Formally, this is the optimization for
\begin{equation}
\argmin_{\mathcal{P}\in \mathcal{P}_n} \sum^N_{i=1}{w_1 \lvert\lvert {_i}F^a_t - \mathcal{P}({_i}F^a_{t+1})\rvert\rvert_2} + w_2 \phi( {_i}B^a_t - \mathcal{P}({_i}B^a_{t+1}))
    \label{eqn:hungarian}
\end{equation}
where $N$ is the total number of nodes in the graphs $G^a_t$ and $G^a_{t+1}$, $\mathcal{P}_n$ is the space over all permutation matrices, $\phi$ is the Intersection over Union (IoU) distance between two bounding boxes, the function $\mathcal{P}(\cdot)$ results in the transformation of a given set of nodes after applying a permutation matrix, i.e., $v \mapsto{\mathcal{P}v}$, and $w_1$ and $w_2$ are scaling factors to balance the two difference distances (i.e., between feature distance and IoU distance across nodes, respectively). Finally, based on this learned permutation matrix, we establish temporal edges between the nodes registered across time. The composite adjacency matrix $\mathcal{A}_G$ and the corresponding action feature matrix $\mathcal{F}_G$ are used to construct the spatial-temporal graph ($\mathcal{G}_a$), representing the entire video $\mathcal{V}{=}I_1, I_2,\ldots I_T$. A temporal smoothing is performed on $\mathcal{G}_a$ to get the final actor-level features, as defined by 
\begin{equation}
    \hat{\mathcal{F}}_G = \sigma(\mathcal{A}_G {\mathcal{F}_G} W_t)
    \label{eqn:temporalSmoothing}
\end{equation}
where $W_t$ is the set of learnable parameters for a GCN layer. 
Similar to the spatial smoothing process, this process can be repeated, but empirically, we find one layer is ideal for our experiments. As seen in Section~\ref{sec:results}, temporal smoothing provides substantial gains in group activity recognition and action detection.

\subsection{Social Modeling with Multi-actor Predictive Learning}
In addition to the global, event-level predictive learning introduced in Equation~\ref{eqn:global_pred}, we introduce the notion of multi-actor predictive learning. This allows us to model the spatial-temporal dynamics of all actors, conditioned on their social interactions and the overall event dynamics of the scene. We model this using a multi-actor prediction loss given by
\begin{equation}
\mathcal{L}_{actor} = \frac{1}{N}\sum_{i=0}^N \lvert\lvert _i\hat{F}^a_{t} + \mathcal{P}(_i\hat{F}^a_{t}) \rvert\rvert_2 + \lvert\lvert _iB^a_t + \mathcal{P}(_iB^a_{t+1}) \rvert\rvert_2
    \label{eqn:multiactor_loss}
\end{equation}
where the first term minimizes the differences between the anticipated actor-level features and the actual actor-level features between consecutive frames, and the second minimizes their respective geometry. We anticipate the future feature and geometry of each actor using two fully connected neural networks defined by $_iF^a_{t+1}{=}{W_{act}}*_iF^a_{t}$ and $_iB^a_{t+1}{=}{W_{bb}}*_iF^a_{t}$, respectively. This allows us to train our overall spatial-temporal prediction stack (defined in Equations \ref{eqn:global_pred} and \ref{eqn:multiactor_loss}) and the smoothing layers (Equations \ref{eqn:spatial_msgPassing} and \ref{eqn:temporalSmoothing}) by minimizing the overall prediction errors given by
\begin{equation}
\mathcal{L}_{total} = \lambda_1 \mathcal{L}_{global} + \lambda_2 \mathcal{L}_{actor}
    \label{eqn:overallLoss}
\end{equation}
where $\lambda_1$ and $\lambda_2$ allow us to balance the global event-level prediction loss and the actor-level multi-actor prediction loss. 

\textbf{Inferring Labels.} For group activity recognition, we do mean average pooling over all actor-level features $\mathcal{\hat{F}}_G$ defined in the composite spatial-temporal action graph $\mathcal{G}_a$. K-means clustering is performed on the mean-pooled features to obtain the final labels. K-means clustering over actor-level features $\mathcal{\hat{F}}_G$ provides actor-level action labels. Following prior work~\cite{aakur2019perceptual,aakur2020action,aakur2022actor} Hungarian matching is performed between the predicted labels and groundtruth labels to compute the quantitative metrics, as defined in Section~\ref{sec:results}. A Spectral Clustering model is fit on the adjacency matrix $\mathcal{A}_G$ to find social communities for social activity recognition, following the protocol from the prior work~\cite{ehsanpour2020joint}.

\section{Experimental Evaluation}\label{sec:results}

\textbf{Data.} We use the Collective Activities Dataset (CAD) \cite{choi2009they} and its annotations-augmented version, SocialCAD~\cite{ehsanpour2020joint}, to evaluate our framework. CAD consists of 44 videos taken in unconstrained real-world scenarios of people performing 6 individual actions across 5 group activities. SocialCAD augmented CAD with additional information, such as social group identification of each person and their collective social activity. We follow prior work~\cite{ehsanpour2020joint} and use 31 videos for training and 11 for evaluation. To evaluate the generalization capabilities of our framework to arbitrary action localization, we use three publicly available benchmarks - UCF Sports \cite{soomro2015action}, JHMDB \cite{Jhuang:ICCV:2013}, and THUMOS'13 \cite{jiang2014thumos}. Each dataset contains a varying number of actions (10 in UCF Sports, 21 in JHMDB, and 24 in THUMOS'13) across different domains (sports and daily activities). Each dataset offers a unique challenge for action localization, such as cluttered scenes, highly similar action classes, large camera motion, and object occlusion. We follow prior work~\cite{aakur2022actor,soomro2017unsupervised} and use official train-test splits for all datasets.

\begin{table}[t]
\centering
\caption{\textbf{Group Activity Recognition} results evaluated on the Collective Activities dataset. Accuracy is reported for group activity recognition and mAP for individual action detection. 
Note: ``-'' indicates the model does not \textit{detect} individual actions.
}
\label{table:CAD_SOTA}
\resizebox{.9\textwidth}{!}{
\begin{tabular}{lcccccc}
\toprule
\multirow{2}{*}{Approach} & \multicolumn{3}{c}{Training Requirements} & \multirow{2}{*}{\shortstack{Bbox\\for eval}} & \multirow{2}{*}{\shortstack{Group Activity\\(Acc.)}} & \multirow{2}{*}{\shortstack{Indiv.Action\\(mAP)}} \\
\cmidrule(lr){2-4}
                          & Bboxes & Actor Labels & Grp.Labels &                           &                       &                      \\
\midrule
HDTM\cite{kim2023self} & \ding{51} & \ding{51} & \ding{51} & \ding{51} & 81.5 & -\\
HANs+HCNs\cite{kong2018hierarchical} & \ding{51} & \ding{51} & \ding{51} & \ding{51} & 84.3 & -\\
CCGL\cite{tang2019coherence} & \ding{51} & \ding{51} & \ding{51} & \ding{51} & 90.0 & -\\
CERN \cite{shu2017cern} & \ding{51} & \ding{51} & \ding{51} & \ding{51} & 87.2 & -\\
stagNet \cite{qi2019stagnet} & \ding{51} & \ding{51} & \ding{51} & \ding{51} & 89.1 & -\\
GAIM \cite{kong2022spatio} & \ding{51} & \ding{51} & \ding{51} & \ding{51} & 90.6 & -\\
AT \cite{gavrilyuk2020actor} & \ding{51} & \ding{51} & \ding{51} & \ding{51} & \underline{90.8} & -\\
GroupFormer \cite{li2021groupformer} & \ding{51} & \ding{51} & \ding{51} & \ding{51} & \textbf{93.6} & -\\

\midrule
HIGCIN \cite{yan2020higcin} & \ding{51} & \ding{55} & \ding{51} & \ding{51} & \text{92.5} & -\\
CRM \cite{azar2019convolutional} & \ding{51} & \ding{51} & \ding{51} & \ding{55} &83.4
&-\\
SBGAR \cite{li2017sbgar} & \ding{55} & \ding{51} & \ding{51} & \ding{55} &83.7
&-\\
Zhang et al \cite{zhang2019fast} & \ding{51} & \ding{55} & \ding{51} & \ding{55} &83.7
&-\\
ARG \cite{wu2019learning} & \ding{51} & \ding{51} & \ding{51} & \ding{55} & 86.10 & 49.60\\
Ehsanpour et. al.\cite{ehsanpour2020joint} & \ding{51} & \ding{51} & \ding{51} & \ding{55} & \underline{89.40} & \underline{55.90} \\
HGC-Former\cite{tamura2022hunting} & \ding{51} & \ding{51} & \ding{51} & \ding{55} & \textbf{96.50} & \textbf{64.90}\\\midrule
PredLearn($K=K_{GT}$) & \ding{55} & \ding{55} & \ding{55} & \ding{55} & 62.83 & 2.82\\
AC-HPL($K=K_{GT}$) & \ding{55} & \ding{55} & \ding{55} & \ding{55} & \underline{72.20} & \underline{10.68}\\
Ours ($K=K_{GT}$) & \ding{55} & \ding{55} & \ding{55} & \ding{55} & \textbf{75.95} & \textbf{26.75}\\
Ours ($K=K_{OPT}$) & \ding{55} & \ding{55} & \ding{55} & \ding{55} & \textbf{90.41} & \textbf{33.02}\\

\bottomrule
\end{tabular}
}
\end{table}

\textbf{Metrics.} We use different metrics for evaluating the performance on each task. We use the mean multi-class classification accuracy (MCA) for group activity recognition. For individual action detection, we follow prior work~\cite{ehsanpour2020joint} and use the mean average precision (mAP) as the evaluation metric to account for missed and false detections. To evaluate social activity understanding, we use two different metrics - membership accuracy and social activity recognition, as defined in SocialCAD. The former measures the accuracy of recognizing a person's social group in the video. The latter measures the ability to jointly predict a person's membership and the social activity label. We report the video-level mAP at 0.5 IOU threshold for arbitrary action localization, i.e., the bounding box predictions must have 0.5 spatial IOU in at least 50\% of the frames.

\textbf{Baselines.} We compare against a variety of supervised, weakly supervised, and unsupervised learning approaches for both group activity understanding and action localization. The supervised \cite{kim2023self,kong2018hierarchical,tang2019coherence,shu2017cern,qi2019stagnet,kong2022spatio,gavrilyuk2020actor,li2021groupformer} and weakly supervised learning baselines \cite{yan2020higcin,azar2019convolutional,li2017sbgar,wu2019learning,ehsanpour2020joint,tamura2022hunting} provide solid baselines for comparing the representation learning capabilities of our framework. Unsupervised learning approaches, particularly closely related approaches such as AC-HPL~\cite{aakur2022actor} and PredLearn~\cite{aakur2020action}, allow us to benchmark our approach with others trained under the same settings. We use Hungarian matching for all unsupervised learning baselines to align their predictions with the ground truth labels, following prior work~\cite{aakur2020action,aakur2022actor}. Note that all baselines, except AC-HPL and PredLearn are not trained in a streaming fashion and require strong visual encoders pre-trained on large amounts of video data (e.g., I3D \cite{carreira2017quo} on Kinetics \cite{carreira2017quo}) and fine-tuned for a large number of epochs ($>50$). We do not require either and only use DETR \cite{carion2020end} pre-trained on MS-COCO for person detection and train in a streaming fashion, requiring only one pass through all the videos.

\subsection{Group Activity Recognition}
We first evaluate our approach on the group activity \textit{recognition} task, where the goal is to identify the activity in which the \textit{majority} of the people are involved. Table~\ref{table:CAD_SOTA} summarizes the results. As can be seen, we perform competitively with supervised learning approaches and significantly outperform prior unsupervised learning approaches such as PredLearn~\cite{aakur2020action} (by $13.12\%$) and AC-HPL~\cite{aakur2022actor} (by $3.75\%$). We observe that some activity classes, such as ``walking'' and ``crossing'', exhibit high intra-class variation in the clustering. Hence, we increase the number of clusters for recognition to its optimal number (using the elbow method with intra-cluster variation as the metric) and devise a baseline indicated by $K=K_{OPT}$. We observe that the accuracy increases significantly to $90.41\%$, outperforming many of the supervised and weakly supervised approaches. It is to be noted that the supervised learning approaches (at the top of Table~\ref{table:CAD_SOTA}) require ground truth bounding boxes during inference for efficient recognition. Weakly supervised approaches~\cite{wu2019learning,tamura2022hunting,ehsanpour2020joint} do not require bounding boxes during inference but require supervision from dense annotations.
Interestingly, we observe that the mean per-class accuracy (MPCA) is $81.25\%$, with the class ``Waiting'' being the worst-performing one at $35.51\%$. We attribute it to the predictive learning paradigm, which naturally focuses on actors with the least predictive motions. It has actors with highly predictable motion, which reduces the model's attention on them and leads to poorer recognition accuracy. However, other classes have a recognition accuracy above $90\%$, indicating the model's effectiveness in recognizing actions that involve reasoning over actor appearance and motion. 

In addition to group activity recognition, we also report the mAP score for individual action detection (last column of Table~\ref{table:CAD_SOTA}), where the goal is to localize and recognize every actor's actions. As can be seen, we once again outperform prior unsupervised learning approaches significantly while offering competitive performance to supervised learning approaches~\cite{wu2019learning,ehsanpour2020joint,tamura2022hunting}. We do not finetune our ROI prediction (DETR) on the CAD dataset as with the supervised learning approaches. This significantly increases the number of actors detected in the scene, which is not always reflected in the ground truth. One such instance is highlighted in Figure~\ref{fig:qual}, where it can be seen that we correctly localize and recognize the individual actions of \textit{all} actors in the scene and not just those in the ground truth. Note that prior unsupervised learning approaches (PredLearn and AC-HPL) do not predict distinct actions for each actor, but rather, a collective group activity is assigned to each person. This reduces their utility in action detection and stems from their inherent assumption that there is one action per video and that all actors participate in this global action. We do not make such assumptions and hence can effectively recognize and localize multiple, simultaneous actions performed by multiple actors.  

\begin{table}[t]
\centering
\caption{\textbf{Social Activity Understanding} results on the SocialCAD dataset~\cite{ehsanpour2020joint}. Note: All results are in the detection setting, i.e., without GT bounding boxes.}
\label{table:socialCAD}
\resizebox{.9\textwidth}{!}{
\begin{tabular}{lcccccc}
\toprule
\multirow{2}{*}{Approach} & \multicolumn{3}{c}{Training Requirements}  & \multirow{2}{*}{\shortstack{Membership\\Recognition}} & \multirow{2}{*}{\shortstack{Social Activity\\ Recognition}} \\
\cmidrule(lr){2-4}
                          & Bboxes & Labels & Member &                           &                       &                      \\
\midrule
GT [Group] (Upper Bound) & - & - & -& 54.4& 51.6\\
HGC-Former \cite{tamura2022hunting}& \ding{51} & \ding{51} & \ding{51} & - & 46.0\\
ARG [Group] \cite{wu2019learning} & \ding{51} & \ding{51} & \ding{51} & 49.0& 34.8\\
Ehsanpour et al [Group] \cite{ehsanpour2020joint} & \ding{51} & \ding{51} & \ding{55} & 49.0& 35.6 \\
Ours & \ding{55} & \ding{55} & \ding{55} & \textbf{32.33}& \textbf{25.07}\\\bottomrule
\end{tabular}
}
\end{table}

\subsection{Social Activity Understanding}
\begin{figure}[t]
    \centering
    \begin{tabular}{ccc}
    \toprule
    \multicolumn{3}{c}{\textbf{Successful Social Activity Detection}}\\
    \midrule
         \includegraphics[width=0.25\textwidth]{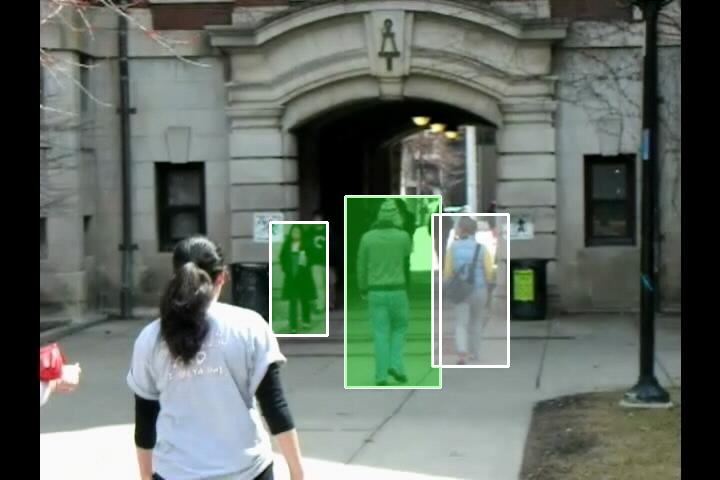} & \includegraphics[width=0.25\textwidth]{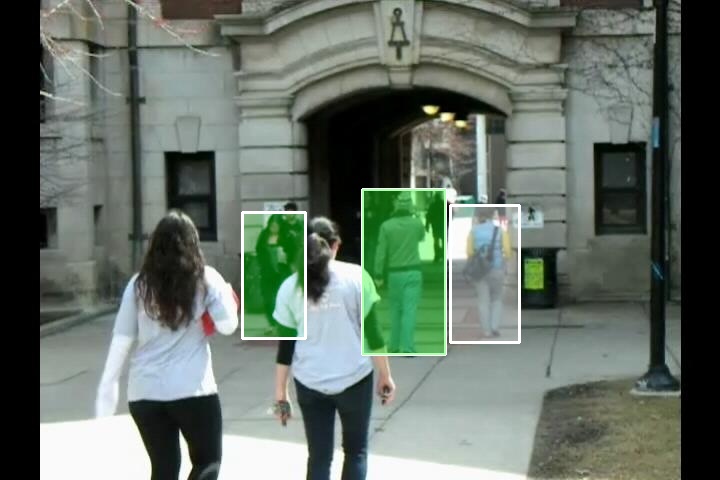} & 
         \includegraphics[width=0.25\textwidth]{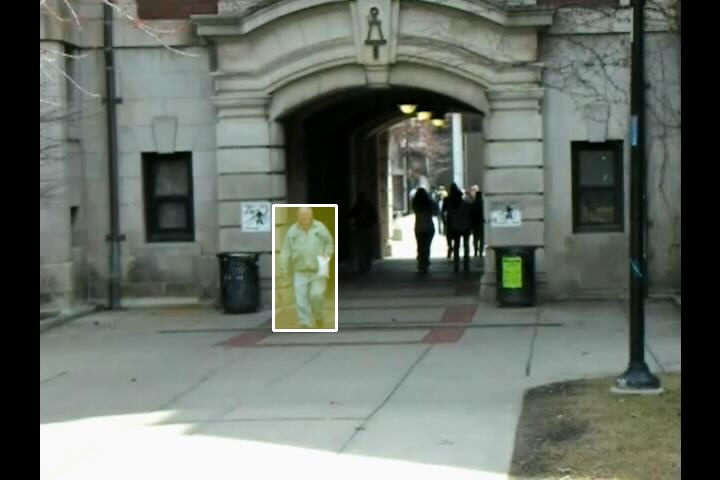}\\
         \includegraphics[width=0.25\textwidth]{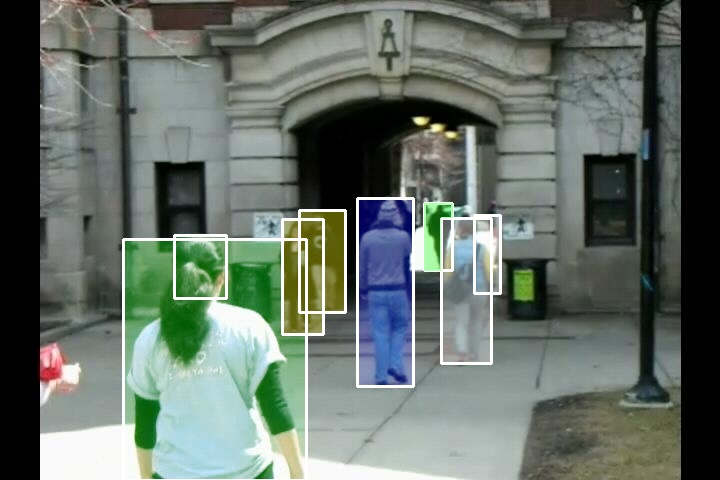} & \includegraphics[width=0.25\textwidth]{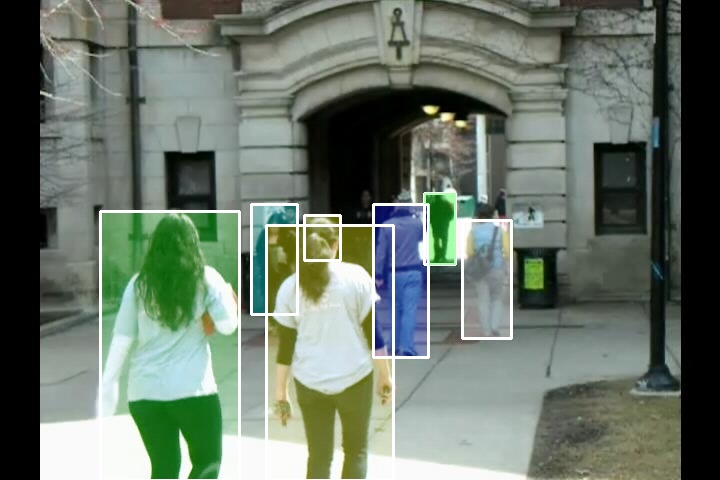} & 
         \includegraphics[width=0.25\textwidth]{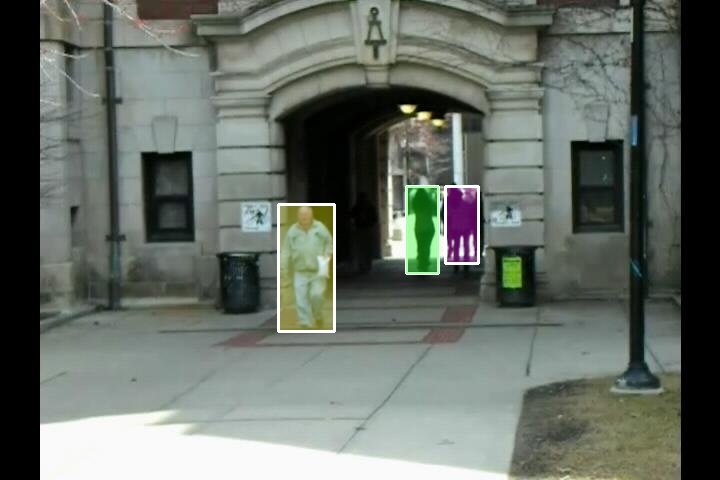}\\
         \midrule
         \multicolumn{3}{c}{\textbf{Unsuccessful Social Activity Detection}}\\
            \midrule
    
             \includegraphics[width=0.25\textwidth]{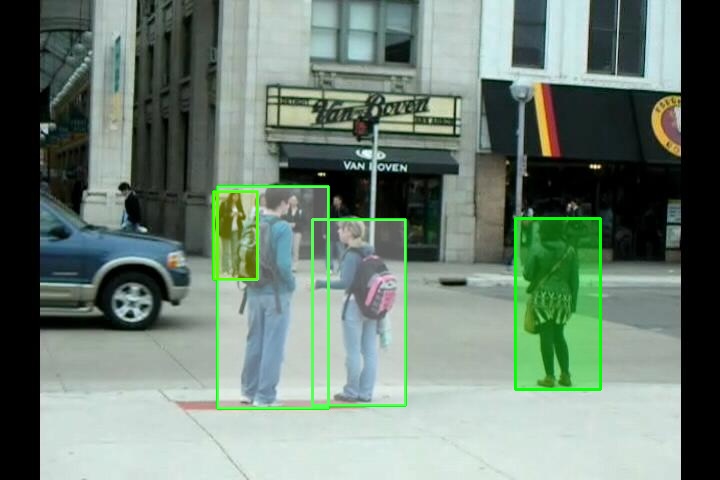} & \includegraphics[width=0.25\textwidth]{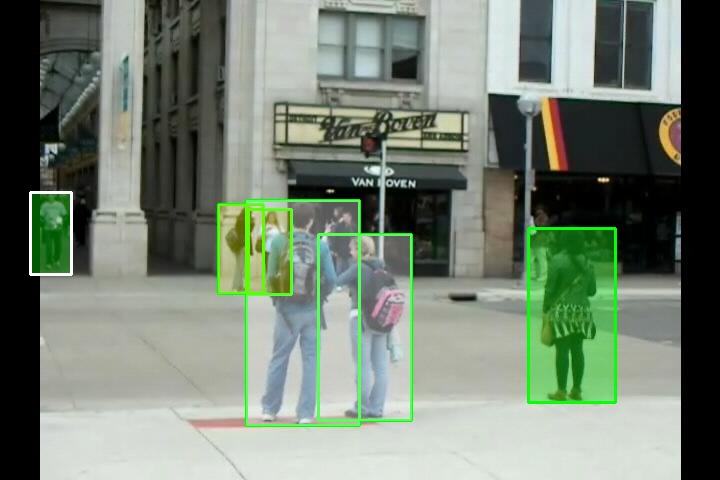} & 
         \includegraphics[width=0.25\textwidth]{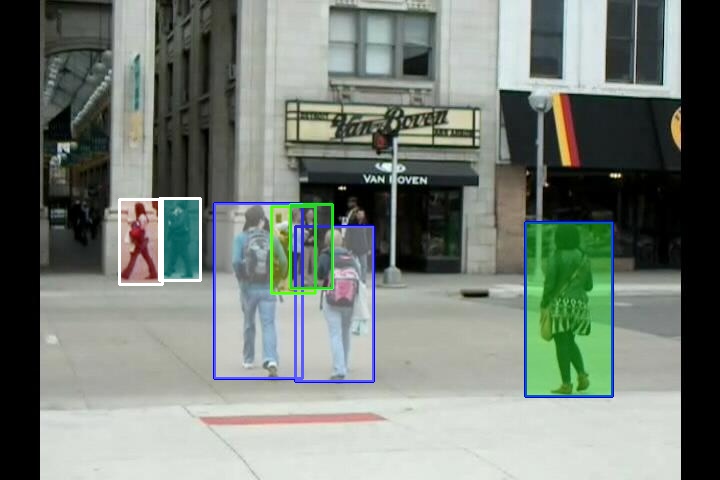}\\
         \includegraphics[width=0.25\textwidth]{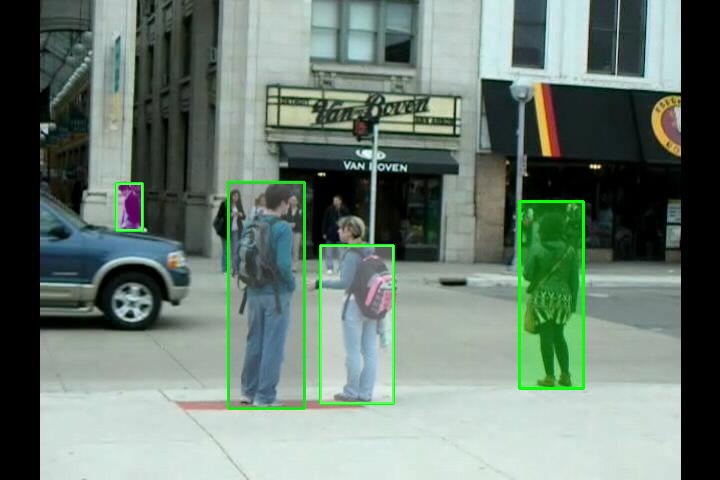} & \includegraphics[width=0.25\textwidth]{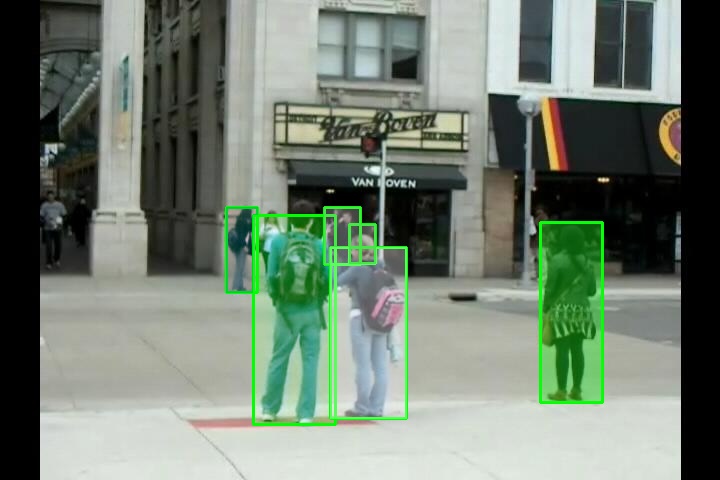} & 
         \includegraphics[width=0.25\textwidth]{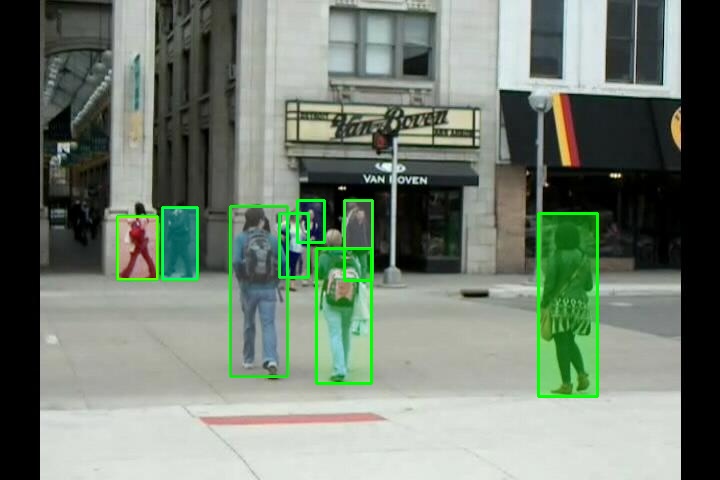}\\
         \bottomrule
    \end{tabular}
    \caption{\textbf{Qualitative visualization} successful (top) and unsuccessful (bottom) activity detection on the Collective Activities dataset. People from the same social group are highlighted in the same color, and the bounding box color indicates their social activity. 
    }
    \label{fig:qual}
\end{figure}

In addition to group activity recognition and individual action detection, we also evaluate the representation learning capabilities of our framework to social activity understanding tasks such as membership recognition and social activity recognition. 
For membership recognition, we follow Ehsanpour \textit{et al.}~\cite{ehsanpour2020joint} and use graph spectral clustering to segment the individual actors into social groups to compute the membership recognition accuracy. 
Table~\ref{table:socialCAD} summarizes the results. 
We perform competitively with supervised learning approaches such as HGC-Former~\cite{tamura2022hunting} and ARG~\cite{wu2019learning}, which require prior knowledge of memberships during training. We also perform competitively with Ehsanpour \textit{et al.}~\cite{ehsanpour2020joint}, which does not require membership labels during training but does require other annotations, such as individual and group labels, along with their bounding box annotations during training. The baselines GT[Group], taken from Ehsanpour \textit{et al.}~\cite{ehsanpour2020joint}, provides the upper bound for detection-based models when the member locations and actions are provided, and an I3D model \cite{carreira2017quo} is used for labeling the membership and social activity of the person. 
On inspecting the results, we find that much of the reduction in membership recognition accuracy is because we predict and localize more actions than provided in the ground truth and, hence, make more predictions per frame. For example, in Figure~\ref{fig:qual}, we detect the membership and actions of all people, not just those in the annotations. We anticipate that fine-tuning DETR on the ground truth annotations and reducing the number of detected people will reduce the false alarms and improve the performance of our approach on these metrics at the cost of generalization. 

\begin{table}[t]
\centering
\caption{\textbf{Generalization to arbitrary action localization.} We report the video-mAP and compare it with unsupervised \textit{action} localization baselines. OOD refers to the evaluation on data other than the training domain (CAD).}
\label{table:OOD_eval}
\begin{tabular}{lcccccc}
\toprule
\multirow{2}{*}{Approach} & \multirow{2}{*}{\shortstack{OOD\\ Eval}}  & \multirow{2}{*}{\shortstack{UCF\\Sport}} & \multirow{2}{*}{\shortstack{JHMDB}} & \multirow{2}{*}{THUMOS13} \\
& & \\
\midrule
Ours & \ding{51}  & \textbf{0.40} & \textbf{0.22} & \textbf{0.15}\\
\midrule
AC-HPL\cite{aakur2022actor} & \ding{55}  & \textbf{0.59} & 0.15 & \underline{0.20}\\
PredLearn\cite{aakur2020action} & \ding{55}  & 0.32 & 0.10 & 0.10\\
Soomro\cite{soomro2017unsupervised} & \ding{55} & 0.30 & \underline{0.22} & 0.06\\    
Ours & \ding{55} & \underline{0.49} & \textbf{0.25} & \textbf{0.21}\\

\bottomrule
\end{tabular}
\end{table}

\textbf{Generalization to Arbitrary Action Localization}
Since our approach does not make any assumptions on the number of actions or type of action, we evaluate its capability to generalize to arbitrary action localization in videos. We evaluate on the UCF Sports \cite{soomro2015action},  JHMDB \cite{Jhuang:ICCV:2013}, and THUMOS'13 \cite{jiang2014thumos} datasets, where there is a single action in the scene with a varying number of actors. Table~\ref{table:OOD_eval} summarizes the results, comparing our approach against other unsupervised learning baselines. We outperform the baselines on all benchmarks, except UCF Sports, when trained on videos from the same domain. The most interesting result is the top row, which shows the performance of our model, trained on CAD and evaluated on out-of-domain videos. We perform well in arbitrary action localization without explicit training, showcasing its generalization capabilities. 

\subsection{Ablation Studies}
We systematically analyze the contributions of each part of our framework and quantify their effects in Table~\ref{table:ablation}. We examine the presence and absence of graph smoothing, temporal smoothing, and the use of action nodes in our action graphs. We see that removing action graphs causes a dramatic decrease in group activity recognition while having minimal effect on individual action recognition. Temporal smoothing has the most impact on both metrics, which could be attributed to the fact that information from the entire video is propagated through the temporal edges and enables better contextualization of group dynamics. Graph smoothing, which enables nodes within the same frame to share information, is essential in propagating information from the action node to each person node. Adding additional layers of temporal and graph smoothing reduces the performance of the approach since it makes the node representations uniform and, hence, loses information about the changes in actor appearances and locations. 

\begin{table}[t]
\caption{\textbf{Ablation study} results on the collective activities dataset. We report accuracy for group activity recognition and the mAP for individual action detection tasks. }
    \centering
    \begin{tabular}{lcc}
        \toprule
         \textbf{Approach} & Group Activity & Indiv. Action\\
         \toprule
         Ours (full model) & 75.95 & 26.75\\
         \midrule
         w/ 2 layers of temporal smoothing & 72.79 & 23.15\\
         w/ 2 layers of graph smoothing & 73.28 & 22.84\\
         w/o temporal smoothing & 59.13 & 14.27\\
         w/o graph smoothing & 68.39 & 11.56\\
         w/o action nodes & 63.46 & 18.28\\
         \bottomrule
         
    \end{tabular}
    \label{table:ablation}
\end{table}

\subsection{Qualitative Evaluation}
Figure~\ref{fig:qual} presents some qualitative visualization of the output from our framework. The top half presents successful social activity detection results. The first row is the ground truth annotations, while the second row shows our corresponding predictions. As can be seen, we can localize and recognize both the social membership (indicated by the color of the shaded region) and the social action (indicated by the bounding box color) of each actor in the scene. Interestingly, we see that we detect and recognize the social activities of people not in the groundtruth (bottom left) and consistently maintain prediction throughout the sequence. The bottom half of Figure~\ref{fig:qual} shows some unsuccessful results where the membership was misclassified, although the social action is correct. We attribute this to our framework's additional action detections that provide ``distractors'' for the membership classification. This effect is also reflected in the individual action mAP score (26.75), where the number of false alarms (due to detections not in the groundtruth annotations) plays a major role. We see that the average recall (across classes) of the groundtruth bounding boxes is $67\%$, indicating that we can recover and label many of the actors correctly. 
\section{Discussion, Limitations, and Future Work}
In this paper, we presented a framework for unsupervised multi-actor, multi-action localization in streaming videos. We showed that it can be adapted to perform group activity recognition, action detection, social membership identification, and social action detection tasks in multi-actor settings. We also demonstrated its potential for localizing an arbitrary number of actions in streaming videos and showed its generalization capabilities by evaluating on out-of-domain data. While it outperformed unsupervised baselines and was competitive with supervised learning approaches, we observe some limitations that offer potential for future work. First, the actor selector module focuses on actions with unpredictable motion. Hence, it fails to consistently localize those with limited predictability, such as ``waiting.'' Similarly, it is sensitive to missed detections. It relies heavily on the ROI detector to provide quality region proposals. Finally, imposing constraints on group formations in frame-level action graphs will likely yield more robust social membership recognition performance. Our future work is focused on improving social action detection by dynamic graph modeling~\cite{bal2022bayesian}.

\textbf{Acknowledgements.} This work was partially supported by the US National Science Foundation Grants IIS 2348689 and IIS 2348690 and the US Department of Agriculture grant 2023-69014-39716-1030191.
\bibliographystyle{splncs04}
\bibliography{references}

\begin{thebibliography}{10}
\providecommand{\url}[1]{\texttt{#1}}
\providecommand{\urlprefix}{URL }
\providecommand{\doi}[1]{https://doi.org/#1}

\bibitem{aakur2020action}
Aakur, S., Sarkar, S.: Action localization through continual predictive learning. In: ECCV. pp. 300--317 (2020)

\bibitem{aakur2022actor}
Aakur, S., Sarkar, S.: Actor-centered representations for action localization in streaming videos. In: ECCV. pp. 70--87 (2022)

\bibitem{aakur2019perceptual}
Aakur, S.N., Sarkar, S.: A perceptual prediction framework for self supervised event segmentation. In: CVPR. pp. 1197--1206 (2019)

\bibitem{azar2019convolutional}
Azar, S.M., Atigh, M.G., Nickabadi, A., Alahi, A.: Convolutional relational machine for group activity recognition. In: CVPR. pp. 7892--7901 (2019)

\bibitem{bal2022bayesian}
Bal, A.B., Mounir, R., Aakur, S., Sarkar, S., Srivastava, A.: Bayesian tracking of video graphs using joint kalman smoothing and registration. In: ECCV. pp. 440--456 (2022)

\bibitem{carion2020end}
Carion, N., Massa, F., Synnaeve, G., Usunier, N., Kirillov, A., Zagoruyko, S.: End-to-end object detection with transformers. In: ECCV. pp. 213--229 (2020)

\bibitem{carreira2017quo}
Carreira, J., Zisserman, A.: Quo vadis, action recognition? a new model and the kinetics dataset. In: CVPR. pp. 6299--6308 (2017)

\bibitem{choi2009they}
Choi, W., Shahid, K., Savarese, S.: What are they doing?: Collective activity classification using spatio-temporal relationship among people. In: ICCV Workshops. pp. 1282--1289 (2009)

\bibitem{ehsanpour2020joint}
Ehsanpour, M., Abedin, A., Saleh, F., Shi, J., Reid, I., Rezatofighi, H.: Joint learning of social groups, individuals action and sub-group activities in videos. In: ECCV. pp. 177--195 (2020)

\bibitem{gavrilyuk2020actor}
Gavrilyuk, K., Sanford, R., Javan, M., Snoek, C.G.: Actor-transformers for group activity recognition. In: CVPR. pp. 839--848 (2020)

\bibitem{han2022dual}
Han, M., Zhang, D.J., Wang, Y., Yan, R., Yao, L., Chang, X., Qiao, Y.: Dual-ai: Dual-path actor interaction learning for group activity recognition. In: CVPR. pp. 2990--2999 (2022)

\bibitem{Jhuang:ICCV:2013}
Jhuang, H., Gall, J., Zuffi, S., Schmid, C., Black, M.J.: Towards understanding action recognition. In: ICCV. pp. 3192--3199 (Dec 2013)

\bibitem{jiang2014thumos}
Jiang, Y.G., Liu, J., Zamir, A.R., Toderici, G., Laptev, I., Shah, M., Sukthankar, R.: Thumos challenge: Action recognition with a large number of classes (2014)

\bibitem{kim2022detector}
Kim, D., Lee, J., Cho, M., Kwak, S.: Detector-free weakly supervised group activity recognition. In: CVPR. pp. 20083--20093 (2022)

\bibitem{kim2023self}
Kim, J., Lee, M., Heo, J.P.: Self-feedback detr for temporal action detection. In: ICCV. pp. 10286--10296 (2023)

\bibitem{kong2022spatio}
Kong, L., Pei, D., He, R., Huang, D., Wang, Y.: Spatio-temporal player relation modeling for tactic recognition in sports videos. IEEE T-CSVT  \textbf{32}(9),  6086--6099 (2022)

\bibitem{kong2018hierarchical}
Kong, L., Qin, J., Huang, D., Wang, Y., Van~Gool, L.: Hierarchical attention and context modeling for group activity recognition. In: ICASSP. pp. 1328--1332 (2018)

\bibitem{li2021groupformer}
Li, S., Cao, Q., Liu, L., Yang, K., Liu, S., Hou, J., Yi, S.: Groupformer: Group activity recognition with clustered spatial-temporal transformer. In: ICCV. pp. 13668--13677 (2021)

\bibitem{li2017sbgar}
Li, X., Choo~Chuah, M.: Sbgar: Semantics based group activity recognition. In: ICCV. pp. 2876--2885 (2017)

\bibitem{lin2014microsoft}
Lin, T.Y., Maire, M., Belongie, S., Hays, J., Perona, P., Ramanan, D., Doll{\'a}r, P., Zitnick, C.L.: Microsoft coco: Common objects in context. In: ECCV. pp. 740--755 (2014)

\bibitem{lotter2016deep}
Lotter, W., Kreiman, G., Cox, D.: Deep predictive coding networks for video prediction and unsupervised learning. arXiv preprint arXiv:1605.08104  (2016)

\bibitem{8769904}
Lu, L., Lu, Y., Yu, R., Di, H., Zhang, L., Wang, S.: Gaim: Graph attention interaction model for collective activity recognition. IEEE T-MM  \textbf{22}(2),  524--539 (2020)

\bibitem{mounir2024streamer}
Mounir, R., Vijayaraghavan, S., Sarkar, S.: Streamer: Streaming representation learning and event segmentation in a hierarchical manner. NeurIPS  \textbf{36} (2024)

\bibitem{pramono2020empowering}
Pramono, R.R.A., Chen, Y.T., Fang, W.H.: Empowering relational network by self-attention augmented conditional random fields for group activity recognition. In: ECCV. pp. 71--90 (2020)

\bibitem{qi2019stagnet}
Qi, M., Wang, Y., Qin, J., Li, A., Luo, J., Van~Gool, L.: Stagnet: An attentive semantic rnn for group activity and individual action recognition. IEEE T-CSVT  \textbf{30}(2),  549--565 (2019)

\bibitem{8621027}
Qi, M., Wang, Y., Qin, J., Li, A., Luo, J., Van~Gool, L.: stagnet: An attentive semantic rnn for group activity and individual action recognition. IEEE T-CSVT  \textbf{30}(2),  549--565 (2020)

\bibitem{raviteja2023sogar}
Raviteja~Chappa, N.V., Nguyen, P., Nelson, A.H., Seo, H.S., Li, X., Dobbs, P.D., Luu, K.: Sogar: Self-supervised spatiotemporal attention-based social group activity recognition. arXiv e-prints pp. arXiv--2305 (2023)

\bibitem{shu2017cern}
Shu, T., Todorovic, S., Zhu, S.C.: Cern: confidence-energy recurrent network for group activity recognition. In: CVPR. pp. 5523--5531 (2017)

\bibitem{soomro2017unsupervised}
Soomro, K., Shah, M.: Unsupervised action discovery and localization in videos. In: ICCV. pp. 696--705 (2017)

\bibitem{soomro2015action}
Soomro, K., Zamir, A.R.: Action recognition in realistic sports videos. In: Computer Vision in Sports, pp. 181--208. Springer (2015)

\bibitem{tamura2022hunting}
Tamura, M., Vishwakarma, R., Vennelakanti, R.: Hunting group clues with transformers for social group activity recognition. In: ECCV. pp. 19--35 (2022)

\bibitem{tang2019coherence}
Tang, J., Shu, X., Yan, R., Zhang, L.: Coherence constrained graph lstm for group activity recognition. IEEE T-PAMI  \textbf{44}(2),  636--647 (2019)

\bibitem{tang2019learning}
Tang, Y., Lu, J., Wang, Z., Yang, M., Zhou, J.: Learning semantics-preserving attention and contextual interaction for group activity recognition. IEEE T-IP  \textbf{28}(10),  4997--5012 (2019)

\bibitem{thilakarathne2024group}
Thilakarathne, H., Nibali, A., He, Z., Morgan, S.: Group activity recognition using unreliable tracked pose. arXiv preprint arXiv:2401.03262  (2024)

\bibitem{trehan2022towards}
Trehan, S., Aakur, S.N.: Towards active vision for action localization with reactive control and predictive learning. In: WACV. pp. 783--792 (2022)

\bibitem{vaswani2017attention}
Vaswani, A., Shazeer, N., Parmar, N., Uszkoreit, J., Jones, L., Gomez, A.N., Kaiser, {\L}., Polosukhin, I.: Attention is all you need. NeurIPS  \textbf{30} (2017)

\bibitem{wang2017recurrent}
Wang, M., Ni, B., Yang, X.: Recurrent modeling of interaction context for collective activity recognition. In: CVPR. pp. 3048--3056 (2017)

\bibitem{wu2019learning}
Wu, J., Wang, L., Wang, L., Guo, J., Wu, G.: Learning actor relation graphs for group activity recognition. In: CVPR. pp. 9964--9974 (2019)

\bibitem{wu2022active}
Wu, L., Lang, X., Xiang, Y., Chen, C., Li, Z., Wang, Z.: Active spatial positions based hierarchical relation inference for group activity recognition. IEEE T-CSVT  (2022)

\bibitem{xie2024active}
Xie, Z., Jiao, C., Wu, K., Guo, D., Hong, R.: Active factor graph network for group activity recognition. IEEE T-IP  (2024)

\bibitem{yan2020higcin}
Yan, R., Xie, L., Tang, J., Shu, X., Tian, Q.: Higcin: Hierarchical graph-based cross inference network for group activity recognition. IEEE T-PAMI  \textbf{45}(6),  6955--6968 (2020)

\bibitem{yan2020social}
Yan, R., Xie, L., Tang, J., Shu, X., Tian, Q.: Social adaptive module for weakly-supervised group activity recognition. In: ECCV. pp. 208--224 (2020)

\bibitem{yuan2021learning}
Yuan, H., Ni, D.: Learning visual context for group activity recognition. In: AAAI Conference on Artificial Intelligence. vol.~35, pp. 3261--3269 (2021)

\bibitem{zacks2001perceiving}
Zacks, J.M., Tversky, B., Iyer, G.: Perceiving, remembering, and communicating structure in events. Journal of Experimental Psychology: General  \textbf{130}(1), ~29 (2001)

\bibitem{zhang2019fast}
Zhang, P., Tang, Y., Hu, J.F., Zheng, W.S.: Fast collective activity recognition under weak supervision. IEEE T-IP  \textbf{29},  29--43 (2019)

\bibitem{zhou2022composer}
Zhou, H., Kadav, A., Shamsian, A., Geng, S., Lai, F., Zhao, L., Liu, T., Kapadia, M., Graf, H.P.: Composer: compositional reasoning of group activity in videos with keypoint-only modality. In: ECCV. pp. 249--266 (2022)

\end{thebibliography}

\end{document}